\documentclass[10pt,twocolumn,letterpaper]{article}

\usepackage[pagenumbers]{cvpr} 

\usepackage[accsupp]{axessibility}  

\makeatletter
\@namedef{ver@everyshi.sty}{}
\makeatother
\usepackage{tikz}
\usepackage{comment}
\usepackage{amsmath,amssymb} 
\usepackage{color}

\usepackage{graphicx}

\usepackage{bm}
\usepackage{multirow}

\usepackage[english]{babel}

\usepackage{algorithm} 
\usepackage{algpseudocode} 
\usepackage{booktabs}
\usepackage{caption}
\usepackage{subcaption}
\usepackage{rotating}

\usepackage{verbatim}
\usepackage{gensymb}
\usepackage{wrapfig}

\usepackage{pgffor}

\usepackage{xspace}
\usepackage{enumitem}

\usepackage[pagebackref,breaklinks,colorlinks]{hyperref}

\usepackage[capitalize]{cleveref}
\crefname{section}{Sec.}{Secs.}
\Crefname{section}{Section}{Sections}
\Crefname{table}{Table}{Tables}
\crefname{table}{Tab.}{Tabs.}

\def\cvprPaperID{1132} 
\def\confName{CVPR}
\def\confYear{2023}

\begin{document}

\title{CloSET: Modeling Clothed Humans on Continuous Surface with\\Explicit Template Decomposition}

\author{Hongwen Zhang$^{1}$  Siyou Lin$^{1}$  Ruizhi Shao$^{1}$  Yuxiang Zhang$^{1}$  Zerong Zheng$^{1}$\\
Han Huang$^{2}$ ~ Yandong Guo$^{2}$ ~ Yebin Liu$^{1}$\\
$^1$Tsinghua University \ \ \ $^2$OPPO Research Institute
}

\maketitle

\begin{abstract}
Creating animatable avatars from static scans requires the modeling of clothing deformations in different poses. Existing learning-based methods typically add pose-dependent deformations upon a minimally-clothed mesh template or a learned implicit template, which have limitations in capturing details or hinder end-to-end learning. In this paper, we revisit point-based solutions and propose to decompose explicit garment-related templates and then add pose-dependent wrinkles to them. In this way, the clothing deformations are disentangled such that the pose-dependent wrinkles can be better learned and applied to unseen poses. Additionally, to tackle the seam artifact issues in recent state-of-the-art point-based methods, we propose to learn point features on a body surface, which establishes a continuous and compact feature space to capture the fine-grained and pose-dependent clothing geometry. To facilitate the research in this field, we also introduce a high-quality scan dataset of humans in real-world clothing. Our approach is validated on two existing datasets and our newly introduced dataset, showing better clothing deformation results in unseen poses. The project page with code and dataset can be found at \href{https://zhanghongwen.cn/closet}{https://zhanghongwen.cn/closet}.

\end{abstract}


\section{Introduction}
\label{sec:intro}

Animating 3D clothed humans requires the modeling of pose-dependent deformations in various poses.
The diversity of clothing styles and body poses makes this task extremely challenging.
Traditional methods are based on either simple rigging and skinning~\cite{baran2007pinocchio,feng2015avatar,liu2019neuroskinning} or physics-based simulation~\cite{deformdynamics,guan2012drape,gundogdu2019garnet,patel20tailornet}, which heavily rely on artist efforts or computational resources.
Recent learning-based methods~\cite{saito2021scanimate,chen2021snarf,ma2021scale,ma2021power} resort to modeling the clothing deformation directly from raw scans of clothed humans.
Despite the promising progress, this task is still far from being solved due to the challenges in clothing representations, generalization to unseen poses, and data acquisition, \etc.

\begin{figure}[t]
	\begin{center}
		\includegraphics[width=0.48\textwidth]{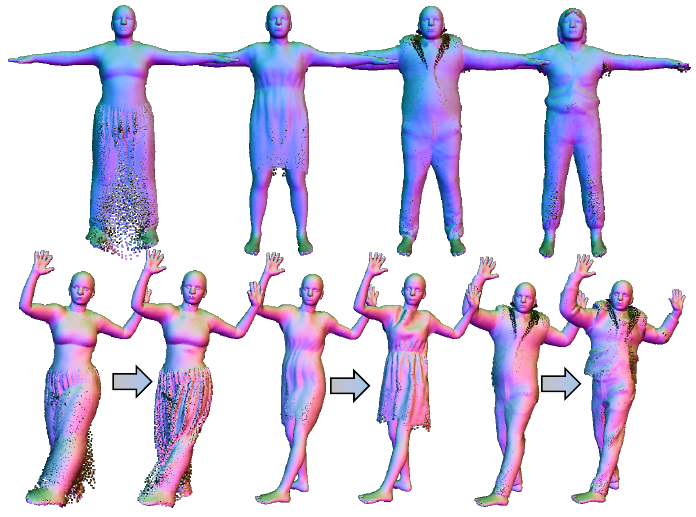}
		\vspace{-5mm}
		\caption{Our method learns to decompose garment templates (top row) and add pose-dependent wrinkles upon them (bottom row).}
		\vspace{-7mm}
		\label{fig:tpl_lbs_posed}
	\end{center}
\end{figure}

For the modeling of pose-dependent garment geometry, the representation of clothing plays a vital role in a learning-based scheme.
As the relationship between body poses and clothing deformations is complex, an effective representation is desirable for neural networks to capture pose-dependent deformations.
In the research of this line, meshes~\cite{alldieck2019tex2shape,ma2020learning,corona2021smplicit}, implicit fields~\cite{saito2021scanimate,chen2021snarf}, and point clouds~\cite{ma2021scale,ma2021power} have been adopted to represent clothing. 
In accordance with the chosen representation, the clothing deformation and geometry features are learned on top of a fixed-resolution template mesh~\cite{ma2020learning,burov2021dynamic}, a 3D implicit sampling space~\cite{saito2021scanimate,chen2021snarf}, or an unfolded UV plane~\cite{alldieck2019tex2shape,corona2021smplicit,ma2021scale,ma2021power}.
Among these representations, the mesh is the most efficient one but is limited to a fixed topology due to its discretization scheme. 
The implicit fields naturally enable continuous feature learning in a resolution-free manner but are too flexible to satisfy the body structure prior, leading to geometry artifacts in unseen poses.
The point clouds enjoy the compact nature and topology flexibility and have shown promising results in the recent state-of-the-art solutions~\cite{ma2021scale,ma2021power} to represent clothing, but the feature learning on UV planes still leads to discontinuity artifacts between body parts.

To model the pose-dependent deformation of clothing, body templates such as SMPL~\cite{loper2015smpl} are typically leveraged to account for articulated motions. However, a body template alone is not ideal, since the body template only models the minimally-clothed humans and may hinder the learning of actual pose-dependent deformations, especially in cases of loose clothing.
To overcome this issue, recent implicit approaches~\cite{saito2021scanimate} make attempts to learn skinning weights in the 3D space to complement the imperfect body templates.
However, their pose-dependent deformations are typically coarse due to the difficulty in learning implicit fields.
For explicit solutions, the recent approach~\cite{lin2022learning} suggests  learning coarse templates implicitly at first and then the pose-dependent deformations explicitly.
Despite its effectiveness, such a workaround requires a two-step modeling procedure and hinders end-to-end learning.

In this work, we propose CloSET, an end-to-end method to tackle the above issues by modeling Clothed humans on a continuous Surface with Explicit Template decomposition.
We follow the spirit of recent state-of-the-art point-based approaches~\cite{ma2021scale,ma2021power,lin2022learning} as they show the efficiency and potential in modeling real-world garments.
We take steps forward in the following aspects for better point-based modeling of clothed humans.
First, we propose to decompose the clothing deformations into explicit garment templates and pose-dependent wrinkles.
Specifically, our method learns a garment-related template and adds the pose-dependent displacement upon them, as shown in Fig.~\ref{fig:tpl_lbs_posed}.
Such a garment-related template preserves a shared topology for various poses and enables better learning of pose-dependent wrinkles.
Different from the recent solution~\cite{lin2022learning} that needs two-step procedures, our method can decompose the explicit templates in an end-to-end manner with more garment details.
Second, we tackle the seam artifact issues that occurred in recent point-based methods~\cite{ma2021scale,ma2021power}.
Instead of using unfolded UV planes, we propose to learn point features on a body surface, which supports a continuous and compact feature space.
We achieve this by learning hierarchical point-based features on top of the body surface and then using barycentric interpolation to sample features continuously.
Compared to feature learning in the UV space~\cite{ma2021power}, on template meshes~\cite{ma2020learning,burov2021dynamic}, or in the 3D implicit space~\cite{saito2019pifu,saito2021scanimate}, our body surface enables the network to capture not only fine-grained details but also long-range part correlations for pose-dependent geometry modeling.
Third, we introduce a new scan dataset of humans in real-world clothing, which contains more than 2,000 high-quality scans of humans in diverse outfits, hoping to facilitate the research in this field.
The main contributions of this work are summarized below:
\begin{itemize}
    \item We propose a point-based clothed human modeling method by decomposing clothing deformations into explicit garment templates and pose-dependent wrinkles in an end-to-end manner.
    These learnable templates provide a garment-aware canonical space so that pose-dependent deformations can be better learned and applied to unseen poses.
    \item We propose to learn point-based clothing features on a continuous body surface, which allows a continuous feature space for fine-grained detail modeling and helps to capture long-range part correlations for pose-dependent geometry modeling.
    \item We introduce a new high-quality scan dataset of clothed humans in real-world clothing to facilitate the research of clothed human modeling and animation from real-world scans.
\end{itemize}

\section{Related Work}\label{sec:related_work}

\paragraph{Representations for Modeling Clothed Humans.}
A key component in modeling clothed humans is the choice of representation, which mainly falls into two categories: implicit and explicit representations.

\emph{Implicit Modeling.} Implicit methods~\cite{park2019deepsdf,mescheder2019occupancy,chibane2020ndf,saito2021scanimate,wang2021metaavatar,chen2021snarf,deng2019neural,mihajlovic2021leap,amoss2020igr,corona2021smplicit,bai2022autoavatar,zheng2021pamir,xiu2022icon,li2022tava} represent surfaces as the level set of an implicit neural scalar field. 
Recent state-of-the-art methods typically learn the clothing deformation field with a canonical space decomposition~\cite{saito2021scanimate,Palafox2021npms,tiwari2021neural,chen2022gdna,li2022avatarcap} or part-based modeling strategies~\cite{deng2020nasa,palafox2022spams,zheng2022structured,qian2022unif,jiang2022lord}.
Compared to mesh templates, implicit surfaces are not topologically constrained to specific templates~\cite{saito2019pifu,saito2020pifuhd}, and can model various clothes with complex topology. However, the learning space of an implicit surface is the whole 3D volume, which makes training and interpolation difficult, especially when the numbers of scan data are limited.

\emph{Explicit Modeling.} Mesh surfaces, the classic explicit representation, currently dominate the field of 3D modeling~\cite{bhatnagar2019mgn,burov2021dynamic,ma2020learning,Neophytou2014layered,tiwari20sizer,yang2018physics,guan2012drape,gundogdu2019garnet,lahner2018deepwrinkles,patel20tailornet,santesteban2019,de2010stable,jiang2020bcnet,vidaurre2020fully,santesteban2021self,bertiche2021deepsd,xiang2021separate,kim2022laplacianfusion} with their compactness and high efficiency in downstream tasks such as rendering, but they are mostly limited to a fixed topology and/or require scan data registered to a template. Thus, mesh-based representations forbid the learning of a universal model for topologically varying clothing types. Though some approaches have been proposed to allow varying mesh topology~\cite{zhu2020deep,pan2019deep,shen2020gan,onizuka2020tetratsdf,wu2020skinning}, they are still limited in their expressiveness. Point clouds enjoy both compactness and topological flexibility. Previous work generates sparse point clouds for 3D representation~\cite{achlioptas2017learning,fan2017point,lin2018learning,zakharkin2021point}. However, the points need to be densely sampled over the surface to model surface geometry accurately. Due to the difficulty of generating a large point set, recent methods group points into patches~\cite{bednarik2020,deng2020better,deprelle2019learning,groueix20183d}. Each patch maps the 2D UV space to the 3D space, allowing arbitrarily dense sampling within this patch. SCALE~\cite{ma2021scale} successfully applies this idea to modeling clothed humans, but produces notable discontinuity artifacts near patch boundaries. POP~\cite{ma2021power} further utilizes a single fine-grained UV map for the whole body surface, leading to a more topologically flexible representation.
However, the discontinuity of the UV map may lead to seam artifacts in POP.
Very recently, FITE~\cite{lin2022learning} suggests learning implicit coarse templates~\cite{zheng2021deep} at first and then explicit fine details.
Despite the efficacy, it requires a two-step modeling procedure.
Concurrently, SkiRT~\cite{ma2022neural} proposes to improve the body template by learning the blend skinning weights with several data terms.
In contrast, our method applies regularization to achieve template decomposition in an end-to-end manner and learns the point-based pose-dependent displacement more effectively.

\paragraph{Pose-dependent Deformations for Animation.}

In the field of character animation, traditional methods utilize rigging and skinning techniques to repose characters~\cite{loper2015smpl,pavlakos2019expressive,baran2007pinocchio,feng2015avatar,liu2019neuroskinning}. but they fail to model realistic pose-dependent clothing deformations such as wrinkles and sliding motions between clothes and body. We conclude two key ingredients in modeling pose-dependent clothing: (i) pose-dependent feature learning; (ii) datasets with realistic clothing.

\emph{Pose-dependent Feature Learning.} Some traditional methods directly incorporate the entire pose parameters into the model~\cite{lahner2018deepwrinkles,ma2020learning,patel20tailornet,yang2018analyzing}. Such methods easily overfit on pose parameters and introduce spurious correlations, causing bad generalization to unseen poses. Recent work explores poses conditioning with local features, either with point clouds~\cite{ma2021scale,ma2021power} or implicit surfaces~\cite{saito2021scanimate,wang2021metaavatar}, and shows superiority in improving geometry quality and in eliminating spurious correlations. Among them, the most relevant to ours is POP~\cite{ma2021power}, which extracts local pose features by utilizing convolution on a UV position map. Despite its compelling performance, POP suffers from artifacts inherent to its UV-based representation, hence the convolution on the UV map produces discontinuity near UV islands' boundaries~\cite{ma2021power}. 
We address this issue by discarding the UV-based scheme and returning to the actual 3D body surface. We attach the features to a uniform set of points on a T-posed body template, and process them via a PointNet++~\cite{qi2017pointnet++} structure for pose-dependent modeling. As validated by our experiments, our pose embedding method leads to both qualitative and quantitative improvements.

\emph{Clothed Human Datasets.} Another challenging issue for training an animatable avatar is the need for datasets of clothed humans in diverse poses. There are considerable efforts seeking to synthesize clothed datasets with physics-based simulation~\cite{deformdynamics,ma2021power,guan2012drape,gundogdu2019garnet,patel20tailornet}. Although they are diverse in poses, there remains an observable domain gap between synthetic clothes and real data. Acquiring clothed scans with realistic details of clothing deformations~\cite{renderpeople,zheng2019deephuman,tao2021function4d,ma2020learning,ponsmoll2017clothcap} is crucial for the development of learning-based methods in this field.

\begin{figure*}[t]
	\begin{center}
		\includegraphics[height=55.5mm]{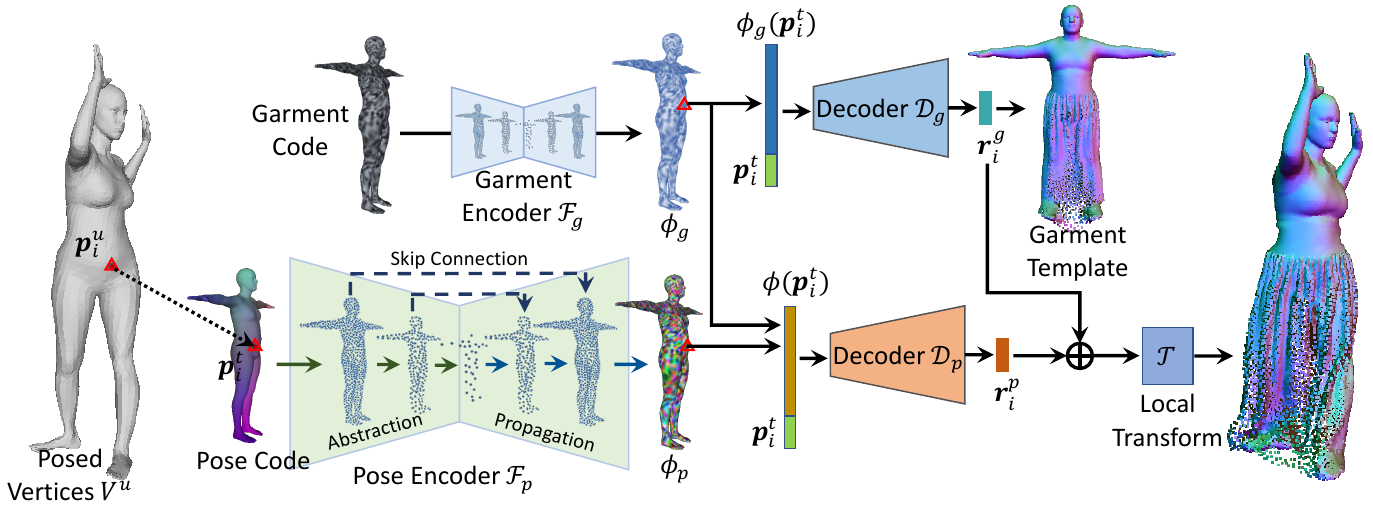}
		\vspace{-2mm}
		\caption{Overview of the proposed method CloSET. Given an input body model, its pose code and garment code are processed hierarchically by point-based pose and garment encoders $\mathcal{F}_p$ and $\mathcal{F}_g$ for the learning of surface features $\bm{\phi}_p$ and $\bm{\phi}_g$.
		For any point $\bm{p}^t_i$ lying on the template surface, its features $\bm{\phi}(\bm{p}^t_i)$ are sampled from surface features accordingly and fed into two decoders for the prediction of the explicit garment template and pose-dependent wrinkle displacements, which will be combined and transformed to the clothing point cloud.}
		\vspace{-5mm}
		\label{fig:framework}
	\end{center}
\end{figure*}

\section{Method}\label{sec:methodology}

As illustrated in Fig.~\ref{fig:framework}, the proposed method CloSET learns garment-related and pose-dependent features on body surfaces (see Sec.~\ref{sec:surf_feat}), which can be sampled in a continuous manner and fed into two decoders for the generation of explicit garment templates and pose-dependent wrinkles (see Sec.~\ref{sec:point_deform}).

\subsection{Continuous Surface Features}
\label{sec:surf_feat}

As most parts of the clothing are deformed smoothly in different poses, a continuous feature space is desirable to model the garment details and pose-dependent garment geometry.
To this end, our approach first learns features on top of a body template surface, \ie, a SMPL~\cite{loper2015smpl} or SMPL-X~\cite{pavlakos2019expressive} model in a T-pose.
Note that these features are not limited to those on template vertices as they can be continuously sampled from the body surface via barycentric interpolation.
Hence, our feature space is more continuous than UV-based spaces~\cite{alldieck2019tex2shape,ma2021power}, while being more compact than 3D implicit feature fields~\cite{saito2021scanimate,wang2021metaavatar}.

To model pose-dependent clothing deformations, the underlying unclothed body model is taken as input to the geometry feature encoder.
For each scan, let $\mathbf{V}^u=\{\bm{v}^u_n\}_{n=1}^{N}$ denote the posed vertex positions of the fitted unclothed body model, where $N=6890$ for SMPL~\cite{loper2015smpl} and $N=10475$ for SMPL-X~\cite{pavlakos2019expressive}.
These posed vertices act as the pose code and will be paired with the template vertices $\mathbf{V}^t=\{\bm{v}^t_n\}_{n=1}^{N}$ of the body model in a T-pose, which shares the same mesh topology with $\mathbf{V}^u$.
These point pairs are processed by the pose encoder $\mathcal{F}_p$ to generate the pose-dependent geometry features $\{\bm{\phi}_p(\bm{v}^t_n) \in \mathbb{R}^{C_p}\}_{n=1}^{N}$ at vertices $\mathbf{V}^t$, \ie,
\begin{equation}
\{\bm{\phi}_p(\bm{v}^t_n) \in \mathbb{R}^{C_p}\}_{n=1}^{N} = \mathcal{F}_p(\mathbf{V}^t, \mathbf{V}^u).
\label{eq:posefeat}
\end{equation}

To learn hierarchical features with different levels of receptive fields, we adopt PointNet++~\cite{qi2017pointnet++} as the architecture of the pose encoder $\mathcal{F}_p$, where vertices $\mathbf{V}^t$ are treated as the input point cloud in the PointNet++ network, while vertices $\mathbf{V}^u$ act as the feature of $\mathbf{V}^t$.
As the template vertices $\mathbf{V}^t$ are constant, the encoder $\mathcal{F}_p$ can focus on the feature learned from the posed vertices $\mathbf{V}^u$.
Moreover, the PointNet++ based $\mathcal{F}_p$ first abstracts features from the template vertices $\mathbf{V}^t$ to sparser points $\{\mathbf{V}^t_l\}_{l=1}^{L}$ at $L$ levels, where the number of $\mathbf{V}^t_l$ decreases with $l$ increasing.
Then, the features at $\{\mathbf{V}^t_l\}_{l=1}^{L}$ are further propagated back to $\mathbf{V}^t$ successively.
In this way, the encoder can capture the long-range part correlations of the pose-dependent deformations.

Similar to POP~\cite{ma2021power}, our method can be trained under multi-outfit or outfit-specific settings. 
When trained with multiple outfits, the pose-dependent deformation should be aware of the outfit type, and hence requires the input of the garment features. 
Specifically, the garment-related features $\{\bm{\phi}_g(\bm{v}^t_n)\in\mathbb{R}^{C_g}\}_{n=1}^{N}$ are also defined on template vertices $\mathbf{V}^t$, which are learned by feeding the garment code $\{\bm{\phi}_{gc}(\bm{v}^t_n)\}_{n=1}^{N}$ to a smaller PointNet++ encoder $\mathcal{F}_g$.
Note that the garment-related features $\bm{\phi}_g$ are shared for each outfit across all poses and optimized during the training.
Since both the pose-dependent and garment-related geometry features are aligned with each other, we denote them as the surface features $\{\bm{\phi}(\bm{v}^t_n)\}_{n=1}^{N}$ for simplicity.
Note that the input of $\bm{\phi}_g(\bm{p}^t_i)$ has no side effect on the results when trained with only one outfit, as the garment features are invariant to the input poses.

\begin{figure}[t]
	\begin{center}
		\includegraphics[width=0.48\textwidth]{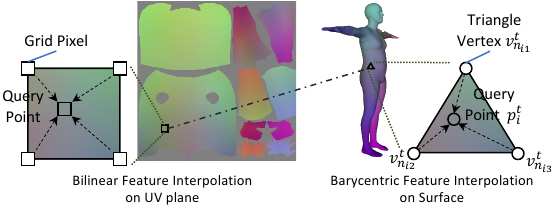}
        \vspace{-6mm}
		\caption{Comparison of the bilinear interpolation on the UV plane and the barycentric interpolation on the surface.}
		\vspace{-6mm}
		\label{fig:bary_sam}
	\end{center}
\end{figure}

\textbf{Continuous Feature Interpolation.}
In the implicit modeling solutions~\cite{saito2019pifu,mihajlovic2021leap,saito2021scanimate}, features are learned in a spatially continuous manner, which contributes to the fine-grained modeling of clothing details.
To sample continuous features in our scheme, we adopt barycentric interpolation on the surface features $\bm{\phi}$.
As illustrated in Fig.~\ref{fig:bary_sam}, for any point $\bm{p}^t_i=\mathbf{V}^t(\bm{b}_i)$ lying on the template surface, where $\bm{b}_i=[n_{i1},n_{i2},n_{i3},b_{i1},b_{i2},b_{i3}]$ denotes the corresponding vertex indices and barycentric coordinates in $\mathbf{V}^t$.
Then, the corresponding surface features can be retrieved via barycentric interpolation, \ie,
\begin{equation}
\bm{\phi}(\bm{p}^t_i) = \sum^{3}_{j=1} ( b_{ij}* \bm{\phi}(\bm{v}^t_{n_{ij}})).
\label{eq:baryfeat}
\end{equation}
In this way, the point features are not limited to those learned on the template vertices and are continuously defined over the whole body surface without the seaming discontinuity issue in the UV plane.

\subsection{Point-based Clothing Deformation}
\label{sec:point_deform}

Following previous work~\cite{ma2021scale,ma2021power}, our approach represents the clothed body as a point cloud.
For any point on the surface of the unclothed body model, the corresponding features are extracted from surface features to predict its displacement and normal vector.

\textbf{Explicit Template Decomposition.}
Instead of predicting the clothing deformation directly, our method decomposes the deformations into two components: garment-related template displacements and pose-dependent wrinkle displacements.
To achieve this, the garment-related template is learned from the garment-related features and shared across all poses.
Meanwhile, the learning of pose-dependent wrinkles are conditioned on both garment-related and pose-dependent features.
Specifically, for the point $\bm{p}^u_i=\mathbf{V}^u(\bm{b}_i)$ at the unclothed body mesh, it has the same vertex indices and barycentric coordinates $\bm{b}_i$ as the point $\bm{p}^t_i$ on the template surface.
The pose-dependent and garment-related features of the point $\bm{p}^u_i$ are first sampled according to Eq.~\eqref{eq:baryfeat} based on $\bm{p}^t_i=\mathbf{V}^t(\bm{b}_i)$ and then further fed into the garment decoder $\mathcal{D}_g$ and the pose decoder $\mathcal{D}_p$ for displacement predictions, \ie,
\begin{equation}
\begin{aligned}
\bm{r}_i^g &= \mathcal{D}_g(\bm{\phi}_g(\bm{p}^t_i), \bm{p}^t_i),\\
\bm{r}_i^p &= \mathcal{D}_p(\oplus(\bm{\phi}_g(\bm{p}^t_i), \bm{\phi}_p(\bm{p}^t_i)), \bm{p}^t_i),
\end{aligned}
\label{eq:displacement}
\end{equation}
where $\oplus$ denotes the concatenation operation, $\bm{r}_i^g$ and $\bm{r}_i^p$ are the displacements for garment templates and pose-dependent wrinkles, respectively.
Finally, $\bm{r}_i^g$ and $\bm{r}_i^p$ will be added together as the clothing deformation $\bm{r}_i=\bm{r}_i^g + \bm{r}_i^p$.

\textbf{Local Transformation.}
Similar to~\cite{ma2021scale,ma2021power}, the displacement $\bm{r}_i$ is learned in a local coordinate system.
It is further transformed to the world coordinate system by applying the following transformation, \ie, $\bm{x}_i = \mathcal{T}_i \bm{r}_i + \bm{p}^u_i$,
where $\mathcal{T}_i$ denotes the local transformation calculated based on the unclothed body model.
Following~\cite{ma2021scale,ma2021power}, the transformation matrix $\mathcal{T}_i$ is defined at the point $\bm{p}^u_i$ on the unclothed body model, which naturally supports the barycentric interpolation.
Similarly, the normal $\bm{n}_i$ of each point is predicted together with $\bm{r}_i$ from $\mathcal{D}$ and transformed by $\mathcal{T}_i$.

\subsection{Loss Functions}

Following previous work~\cite{ma2021scale,ma2021power}, the point-based clothing deformation is learned with the summation of loss functions: $\mathcal{L} = \mathcal{L}_{data} + \lambda_{rgl}\mathcal{L}_{rgl}$.
where $\mathcal{L}_{data}$ and $\mathcal{L}_{rgl}$ denote the data and regularization terms respectively, and the weight $\lambda_{rgl}$ balances the loss terms.

\textbf{Data Term.}
The data term $\mathcal{L}_{data}$ is calculated on the final predicted points and normals, \ie, $\mathcal{L}_{data} = \lambda_{p}\mathcal{L}_{p} + \lambda_{n}\mathcal{L}_{n}$.
Specifically, $\mathcal{L}_{p}$ is the normalized Chamfer distance to minimize the bi-directional distances between the point sets of the prediction and the ground-truth scan: \newline $\mathcal{L}_{p}=Chamfer\left(\{\bm{x}_i\}_{i=1}^{M},\{\hat{\bm{x}}_j\}_{j=1}^{N_s}\right)=$
\begin{equation}
\frac{1}{M} \sum_{i=1}^{M} \min _{j}\|\bm{x}_{i}-\hat{\bm{x}}_j\|_2^{2} +\frac{1}{N_s}\sum_{j=1}^{N_s} \min_{i}\|\bm{x}_{i}-\hat{\bm{x}}_j\|_2^{2},
\label{eq:chamfer_loss}
\end{equation}
where $\hat{\bm{x}}_j$ is the point sampled from the ground-truth surface, $M$ and $N_s$ denote the number of the predicted and ground-truth points, respectively.

The normal loss $\mathcal{L}_{n}$ is an averaged $L1$ distance between the normal of each predicted point and its nearest ground-truth counterpart:
\begin{equation}
\mathcal{L}_{n} =L1\left(\{\bm{n}_i\}_{i=1}^{M},\{\hat{\bm{n}}_i\}_{i=1}^{M}\right) = \frac{1}{M}\sum_{i=1}^M \|\bm{n}_i-\hat{\bm{n}}_i\|,
\label{eq:norm_loss}
\end{equation}
where $\hat{\bm{n}}_i$ is the normal of its nearest point in the ground-truth point set.

Note that we do not apply data terms on the garment templates, as we found such a strategy leads to noisy template learning in our experiments.

\textbf{Regularization Term.}
The regularization terms are added to prevent the predicted deformations from being extremely large and regularize the garment code. Moreover, following the previous implicit template learning solution~\cite{zheng2021deep,li2022avatarcap}, we also add regularization on the pose-dependent displacement $\bm{r}_i^p$ to encourage it to be as small as possible. 
As the pose-dependent displacement represents the clothing deformation in various poses, such a regularization implies that the pose-invariant deformation should be retained in the template displacement $\bm{r}_i^g$, which forms the garment-related template shared by all poses.
Overall, the regularization term can be written as follows:
\begin{equation}
\mathcal{L}_{rgl} = \frac{1}{M}\sum_{i=1}^{M}\|\mathbf{r}_{i}\|_2^2  + \frac{\lambda_{pd}}{M}\sum_{i=1}^{M}\|\mathbf{r}_{i}^p\|_2^2 +  \frac{\lambda_{gc}}{N}\sum_{n=1}^{N}\|\bm{\phi}_{gc}(\bm{v}^t_n)\|_2^2.
\end{equation}

\section{Experiments}
\label{sec:experiments}

\paragraph{Network Architecture.}
For a fair comparison with POP~\cite{ma2021power}, we modify the official PointNet++~\cite{qi2017pointnet++} (PN++) architecture so that our encoders have comparable network parameters as POP~\cite{ma2021power}.
The modified PointNet++ architecture has 6 layers for feature abstraction and 6 layers for feature propagation (\ie, $L=6$).
Since the input point cloud $\mathbf{V}^t$ has constant coordinates, the farthest point sampling in PointNet++ is only performed at the first forward process, and the sampling indices are saved for the next run.
In this way, the runtime is significantly reduced for both training and inference so that our pose and garment encoders can have similar network parameters and runtime speeds to POP.
Note that the pose and garment encoders in our method can also be replaced with recent state-of-the-art point-based encoders such as PointMLP~\cite{ma2022rethinking} and PointNeXt~\cite{qian2022pointnext}.
More details about the network architecture and implementation can be found in the Supp.Mat.

\paragraph{Datasets.}
We use CAPE~\cite{ma2020learning}, ReSynth~\cite{ma2021power}, and our newly introduced dataset THuman-CloSET for training and evaluation.
\newline
\textbf{CAPE}~\cite{ma2020learning} is a captured human dataset consisting of multiple humans in various motions.
The outfits in this dataset mainly include common clothing such as T-shirts.
We follow SCALE~\cite{ma2021scale} to choose blazerlong (with outfits of blazer jacket and long trousers) and shortlong (with outfits of short T-shirt and long trousers) from subject 03375 to validate the efficacy of our method.
\newline
\textbf{ReSynth}~\cite{ma2021power} is a synthetic dataset introduced in POP~\cite{ma2021power}. It is created by using physics simulation, and contains challenging outfits such as skirts and jackets.
We use the official training and test split as~\cite{ma2021power}.
\newline
\textbf{THuman-CloSET} is our newly introduced dataset, containing high-quality clothed human scans captured by a dense camera rig. 
We introduce THuman-CloSET for the reason that existing pose-dependent clothing datasets~\cite{ma2020learning,ma2021power} are with either relatively tight clothing or synthetic clothing via physics simulation.
In THuman-CloSET, there are more than 2,000 scans of 15 outfits with a large variation in clothing style, including T-shirts, pants, skirts, dresses, jackets, and coats, to name a few.
For each outfit, the subject is guided to perform different poses by imitating the poses in CAPE.
Moreover, each subject has a scan with minimal clothing in A-pose.
THuman-CloSET contains well-fitted body models in the form of SMPL-X~\cite{pavlakos2019expressive}.
Note that the loose clothing makes the fitting of the underlying body models quite challenging.
For more accurate fitting of the body models, we first fit a SMPL-X model on the scan of the subject in minimal clothing and then adopt its shape parameters for fitting the outfit scans in different poses.
More details can be found in the Supp.Mat.
In our experiments, we use the outfit scans in 100 different poses for training and use the remaining poses for evaluation.
We hope our new dataset can open a promising direction for clothed human modeling and animation from real-world scans.

\paragraph{Metrics.}
Following previous work~\cite{ma2021power}, we generate 50K points from our method and point-based baselines and adopt the Chamfer Distance (see Eq.~\eqref{eq:chamfer_loss}) and the $L1$ normal discrepancy (see Eq.~\eqref{eq:norm_loss}) for quantitative evaluation.
By default, the Chamfer distance (CD) and normal discrepancy (NML) are reported in the unit of $\times 10^{-4}m^2$ and $\times 10^{-1}$, respectively.
To evaluate the implicit modeling methods, the points are sampled from the surface extracted using Marching Cubes~\cite{lorensen1987marching}.

\begin{table}[t]
  \centering
  \small
  \caption{Quantitative comparison with previous point-based methods on ReSynth. $\dagger$ denotes the methods using 1/8 training data.}
  \vspace{-2mm}
    \begin{tabular}{lcccc}
    \toprule
    \multicolumn{1}{c}{\multirow{2}[4]{*}{Method}} & \multicolumn{2}{c}{Chamfer-$L_2$ $\downarrow$} & \multicolumn{2}{c}{Normal diff. $\downarrow$} \\
\cmidrule{2-5}          & Mean  & Max   & Mean  & Max \\
    \midrule
    \textbf{outfit-specific} \\
    SCALE~\cite{ma2021scale} & 1.491 & 8.451 & 1.041 & 1.321 \\
    POP~\cite{ma2021power} & 1.356 & 7.339 & \textbf{1.013} & \textbf{1.289} \\
    \midrule
    \textbf{multi-outfit} \\
    Baseline (POP~\cite{ma2021power}) $\dagger$ & 1.490 & 7.859 & 1.050 & 1.326 \\
    Baseline w. PN++ $\dagger$ & 1.290 & 5.940 & 1.028 & 1.330 \\
    CloSET (Ours) $\dagger$ & \textbf{1.240} & \textbf{5.543} & 1.019 & 1.315 \\
    \bottomrule
    \end{tabular}%
\vspace{-1mm}
  \label{tab:resyn_multi}%
\end{table}%

\addtolength{\tabcolsep}{-4pt}
\begin{table}[t]
  \centering
  \small
  \caption{Quantitative comparison of different methods on the proposed THuman-CloSET dataset in the outfit-specific setting.}
  \vspace{-2mm}
    \begin{tabular}{l|cc|cc|cc|cc}
    \toprule
    \multicolumn{1}{c}{\multirow{2}[4]{*}{Subject ID}} & \multicolumn{2}{c}{\footnotesize{SCANimate}} & \multicolumn{2}{c}{SNARF} & \multicolumn{2}{c}{POP} & \multicolumn{2}{c}{CloSET} \\
\cmidrule{2-9}          & CD    & NML   & CD    & NML   & CD    & NML   & CD    & NML \\
    \midrule
    sweater-000 & 1.06  & 1.64  & 7.11  & 2.09  & 0.76  & 1.55  & \textbf{0.68} & \textbf{1.48} \\
    longshirt-001 & 1.42  & 1.85  & 6.66  & 2.21  & 1.54  & 1.83  & \textbf{1.39} & \textbf{1.71} \\
    skirt-005 & 1.93  & 1.74  & 9.39  & 2.31  & 1.66  & 1.43  & \textbf{1.49} & \textbf{1.36} \\
    \bottomrule
    \end{tabular}%
  \label{tab:sota_real_world}%
\end{table}%
\addtolength{\tabcolsep}{4pt}

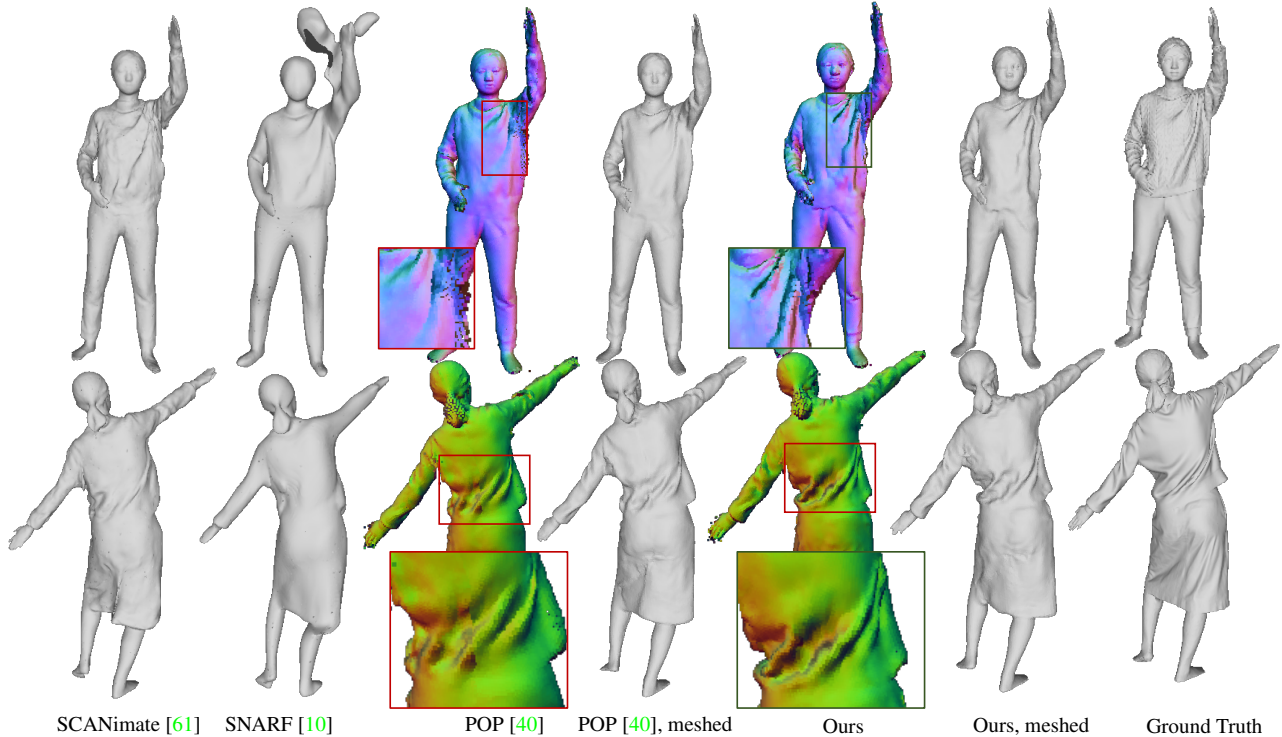
\begin{figure*}[t]
\centering
\includegraphics[width=0.98\textwidth]{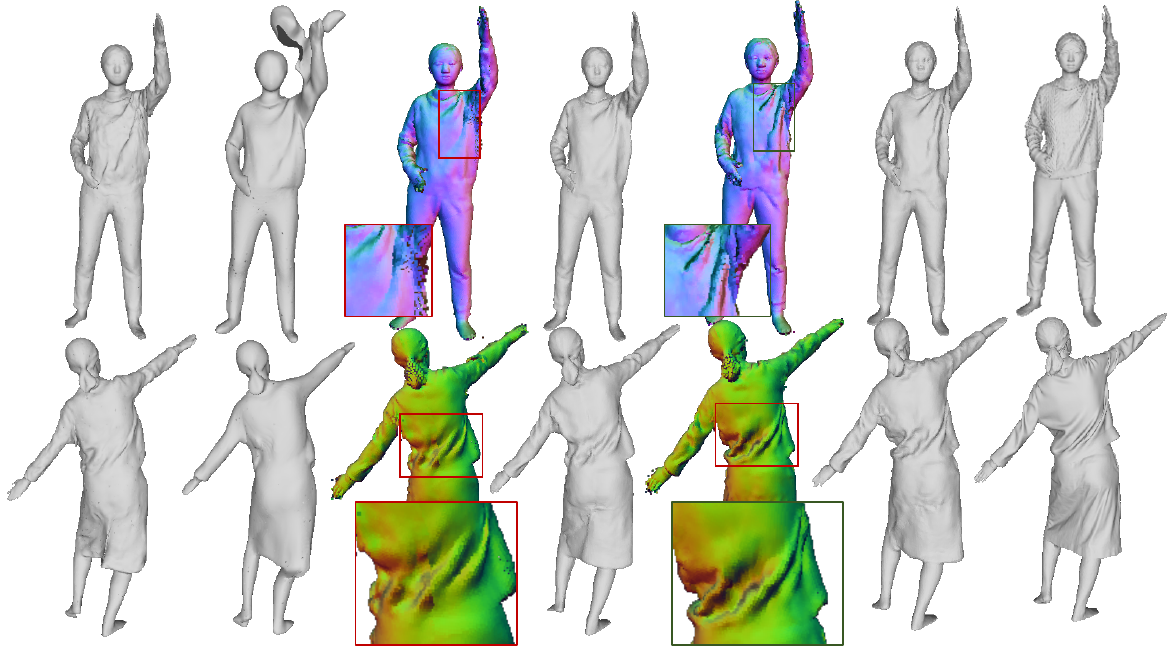}
\begin{tikzpicture}[remember picture,overlay]
	\node[font=\fontsize{8pt}{8pt}\selectfont] at (-15.5,-0.1) {SCANimate~\cite{saito2021scanimate}};
	\node[font=\fontsize{8pt}{8pt}\selectfont] at (-13.5,-0.1) {SNARF~\cite{chen2021snarf}};
 	\node[font=\fontsize{8pt}{8pt}\selectfont] at (-10.5,-0.1) {POP~\cite{ma2021power}};
	\node[font=\fontsize{8pt}{8pt}\selectfont] at (-8.5,-0.1) {POP~\cite{ma2021power}, meshed};
 \node[font=\fontsize{8pt}{8pt}\selectfont] at (-6,-0.1) {Ours};
 	\node[font=\fontsize{8pt}{8pt}\selectfont] at (-3.5,-0.1) {Ours, meshed};
	\node[font=\fontsize{8pt}{8pt}\selectfont] at (-1.2,-0.1) {Ground Truth};
\end{tikzpicture}
\caption{Comparison of different clothed human modeling methods on the proposed real-world scan dataset.}
\vspace{-3mm}
\label{fig:sota_vis}
\end{figure*}

\subsection{Comparison with the State-of-the-art Methods}

We compare results with recent state-of-the-art methods, including point-based approaches SCALE~\cite{ma2021scale}, POP~\cite{ma2021power}, and SkiRT~\cite{ma2022neural}, and implicit approaches SCANimate~\cite{saito2021scanimate} and SNARF~\cite{chen2021snarf}.

\paragraph{ReSynth.}
Tab.~\ref{tab:resyn_multi} reports the results of the pose-dependent clothing predictions on unseen motion sequences from the ReSynth~\cite{ma2021power} dataset, where all 12 outfits are used for evaluation.
As can be seen, the proposed approach has the lowest mean and max errors, which outperforms all other approaches including POP~\cite{ma2021power}.
Note that our approach needs fewer data for the pose-dependent deformation modeling.
By using only $1/8$ data, our approach achieves a performance comparable to or even better than other models trained with full data.
Tab.~\ref{tab:resyn_peroutfit} also reports outfit-specific performances on 3 selected subject-outfit types, including jackets, skirts, and dresses. 
In comparison with the recent state-of-the-art method SkiRT~\cite{ma2022neural},  our method achieves better results on challenging skirt/dress outfits and comparable results on non-skirt clothing.

\addtolength{\tabcolsep}{-4pt}
\begin{table}[t]
  \centering
  \small
  \caption{Quantitative comparison of different methods on the ReSynth dataset in the outfit-specific setting. The garment styles are non-skirt, skirt, and dress for carla-004, christine-027, and felice-004, respectively.}
  \vspace{-2mm}
    \begin{tabular}{l|cc|cc|cc|cc}
    \toprule
    \multicolumn{1}{c}{\multirow{2}[4]{*}{Subject ID}} & \multicolumn{2}{c}{\footnotesize{SCANimate}} & \multicolumn{2}{c}{POP} & \multicolumn{2}{c}{SkiRT} & \multicolumn{2}{c}{CloSET} \\
\cmidrule{2-9}          & CD    & NML   & CD    & NML   & CD    & NML   & CD    & NML \\
    \midrule
    carla-004 & 0.90 & 1.52 & 0.51 & \textbf{1.02} & \textbf{0.48} & 1.06 & 0.49  & 1.04 \\
    christine-027 & 3.21 & 1.66 & 1.72 & 0.97 & 1.54 & 0.99 & \textbf{1.49} & \textbf{0.97} \\
    felice-004 & 20.79 & 2.94 & 7.34 & 1.24 & 6.45 & 1.25 & \textbf{6.01} & \textbf{1.16} \\
    \bottomrule
    \end{tabular}%
  \label{tab:resyn_peroutfit}%
  \vspace{-2mm}
\end{table}%
\addtolength{\tabcolsep}{4pt}

\paragraph{THuman-CloSET.}
The effectiveness of our method is also validated on our real-world THuman-CloSET dataset.
The sparse training poses and loose clothing make this dataset very challenging for clothed human modeling.
Tab.~\ref{tab:sota_real_world} reports the quantitative comparisons of different methods on three representative outfits.
Fig.~\ref{fig:sota_vis} also shows example results of different methods, where we follow previous work~\cite{ma2021power} to obtain meshed results via Poisson surface reconstruction. 
We can see that our method generalizes better to unseen poses and produces more natural pose-dependent wrinkles than other methods.
In our experiments, we found that SNARF~\cite{chen2021snarf} fails to learn correct skinning weights due to loose clothing and limited training poses.
As discussed in FITE~\cite{lin2022learning}, there is an ill-posed issue of jointly optimizing the canonical shape and the skinning fields, which becomes more severe in our dataset.

\subsection{Ablation Study}

\begin{figure}[htb]
\centering
\includegraphics[width=0.5\textwidth]{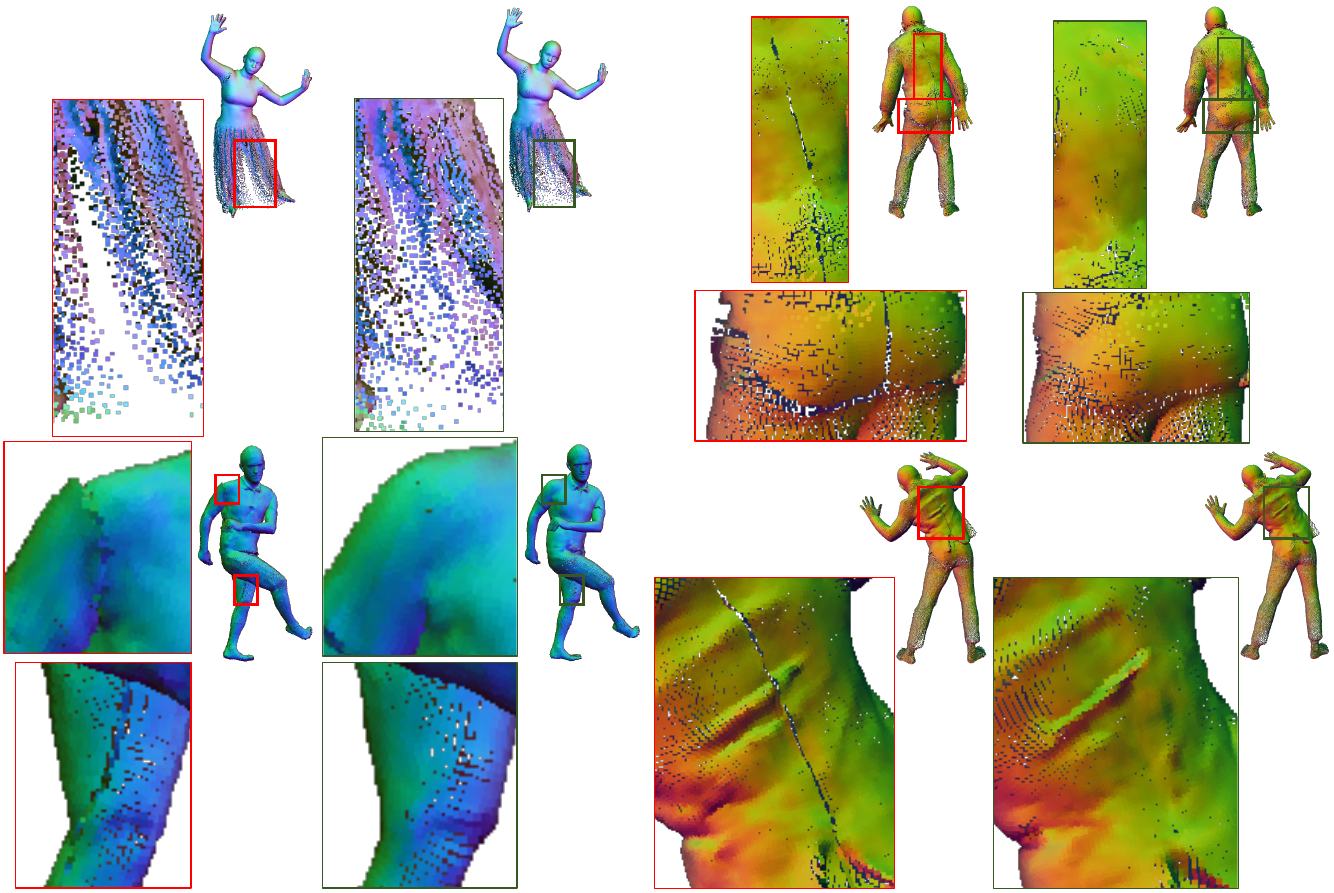}
\begin{tikzpicture}[remember picture,overlay]
	\node[font=\fontsize{8pt}{8pt}\selectfont] at (-3,0.2) {POP~\cite{ma2021power}};
	\node[font=\fontsize{8pt}{8pt}\selectfont] at (-1,0.2) {Ours};
 	\node[font=\fontsize{8pt}{8pt}\selectfont] at (1,0.2) {POP~\cite{ma2021power}};
	\node[font=\fontsize{8pt}{8pt}\selectfont] at (3,0.2) {Ours};
\end{tikzpicture}
\vspace{-3mm}
\caption{Comparison of the approach learned on UV planes (POP~\cite{ma2021power}) and the approach learned on continuous surfaces (Ours). Our solution alleviates the seam artifacts of POP.}
\vspace{-5mm}
\label{fig:cape_res}
\end{figure}
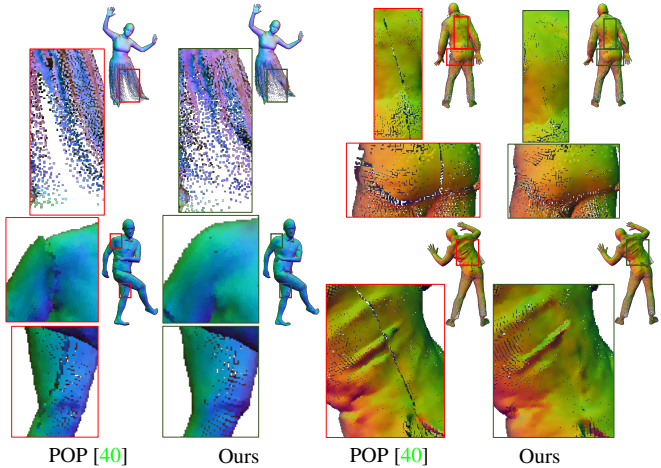

\paragraph{Evaluation of Continuous Surface Features.}
The point features in our method are learned on the body surface, which provides a continuous and compact feature learning space.
To validate this, Tab.~\ref{tab:feat_space} summarizes the feature learning space of different approaches and their performances on two representative outfits from CAPE~\cite{ma2020learning}.
Here, we only include the proposed Continuous Surface Features (CSF) in Tab.~\ref{tab:feat_space} by applying the continuous features on POP~\cite{ma2021power} for fair comparisons with SCALE~\cite{ma2021scale} and POP~\cite{ma2021power}.
As discussed previously, existing solutions learn features either in a discontinuous space (\eg, CAPE~\cite{ma2020learning} on the fixed resolution mesh, POP~\cite{ma2021power} on the 2D UV plane) or in a space that is too flexible (\eg, NASA~\cite{deng2020nasa} in the implicit 3D space), while our approach learns features on continuous and compact surface space.
Though SCALE~\cite{ma2021scale} has also investigated using the point-based encoder (PointNet~\cite{qi2017pointnet}) for pose-dependent feature extraction, it only uses the global features which lack fine-grained information. 
In contrast, we adopt PointNet++~\cite{qi2017pointnet++} (PN++) to learn hierarchical surface features, so that the pose-dependent features can be learned more effectively.
Fig.~\ref{fig:cape_res} shows the qualitative results of the ablation approaches learned on UV planes and continuous surfaces. We can see that our solution clearly alleviates the seam artifacts of POP.

\addtolength{\tabcolsep}{-1pt}
\begin{table}[t]
  \centering
  \scriptsize
  \caption{Comparison of the modeling ability of different approaches and their feature learning space on the CAPE dataset.}
  \vspace{-2mm}
    \begin{tabular}{llcccc}
    \toprule
    \multirow{2}[4]{*}{Methods} & \multicolumn{1}{c}{\multirow{2}[4]{*}{Features}} & \multicolumn{2}{c}{Chamfer-$L_2$ $\downarrow$} & \multicolumn{2}{c}{Normal diff. $\downarrow$} \\
\cmidrule{3-6}          &       & \textit{blazerlong} & \textit{shortlong} & \textit{blazerlong} & \textit{shortlong} \\
    \midrule
    CAPE~\cite{ma2020learning}  & Mesh & 1.96  & 1.37  & 1.28  & 1.15 \\
    NASA~\cite{deng2020nasa}  & 3D space & 1.37  & 0.95  & 1.29  & 1.17 \\
    SCALE~\cite{ma2021scale} & Global & 1.46  & 1.03  & 1.34  & 1.16 \\
    SCALE~\cite{ma2021scale} & UV plane & 1.07  & 0.89  & 1.22  & 1.12 \\
    POP~\cite{ma2021power} & UV plane & 0.78  & 0.57  & 1.29  & 1.24 \\
    \midrule
    CSF & Surface & \textbf{0.71}  & \textbf{0.54}  & \textbf{1.15}  & \textbf{1.09} \\
    \bottomrule
    \end{tabular}%
\vspace{-3mm}
  \label{tab:feat_space}%
\end{table}%
\addtolength{\tabcolsep}{1pt}

\begin{figure*}[th]
\centering
\includegraphics[width=0.98\textwidth]{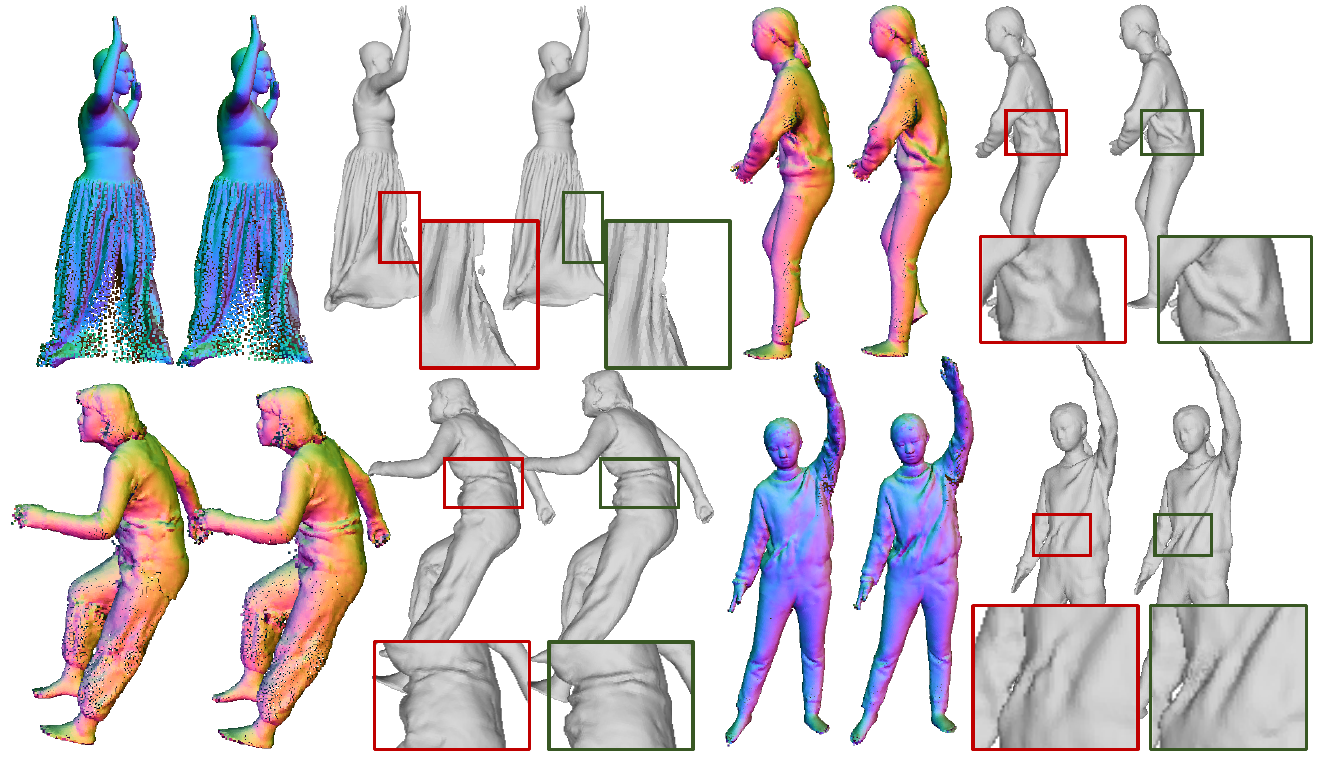}
\begin{tikzpicture}[remember picture,overlay]
	\node[font=\fontsize{8pt}{8pt}\selectfont] at (-16,-0.2) {POP};
	\node[font=\fontsize{8pt}{8pt}\selectfont] at (-14,-0.2) {POP+ETD};
 	\node[font=\fontsize{8pt}{8pt}\selectfont] at (-11.5,-0.2) {POP, meshed};
	\node[font=\fontsize{8pt}{8pt}\selectfont] at (-9.,-0.2) {POP+ETD, meshed};
 \node[font=\fontsize{8pt}{8pt}\selectfont] at (-7,-0.2) {POP};
	\node[font=\fontsize{8pt}{8pt}\selectfont] at (-5.5,-0.2) {POP+ETD};
 	\node[font=\fontsize{8pt}{8pt}\selectfont] at (-3.5,-0.2) {POP, meshed};
	\node[font=\fontsize{8pt}{8pt}\selectfont] at (-1.2,-0.2) {POP+ETD, meshed};
\end{tikzpicture}
\vspace{2mm}
\caption{Comparison of the clothing deformation in unseen poses. Explicit template decomposition (ETD) helps to capture more natural pose-dependent wrinkle details than POP~\cite{ma2021power}.}
\vspace{-5mm}
\label{fig:ablation_etd}
\end{figure*}
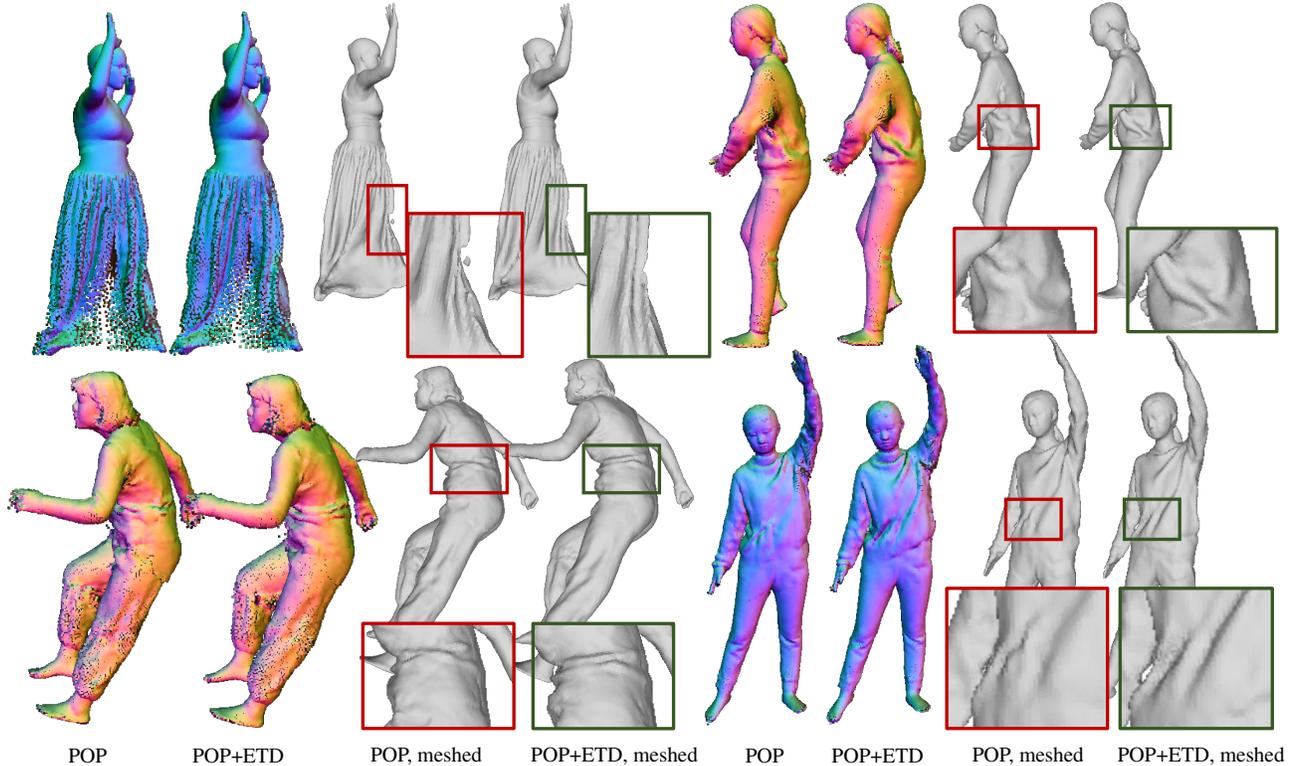

\begin{figure}[th]
\centering
\includegraphics[width=0.5\textwidth]{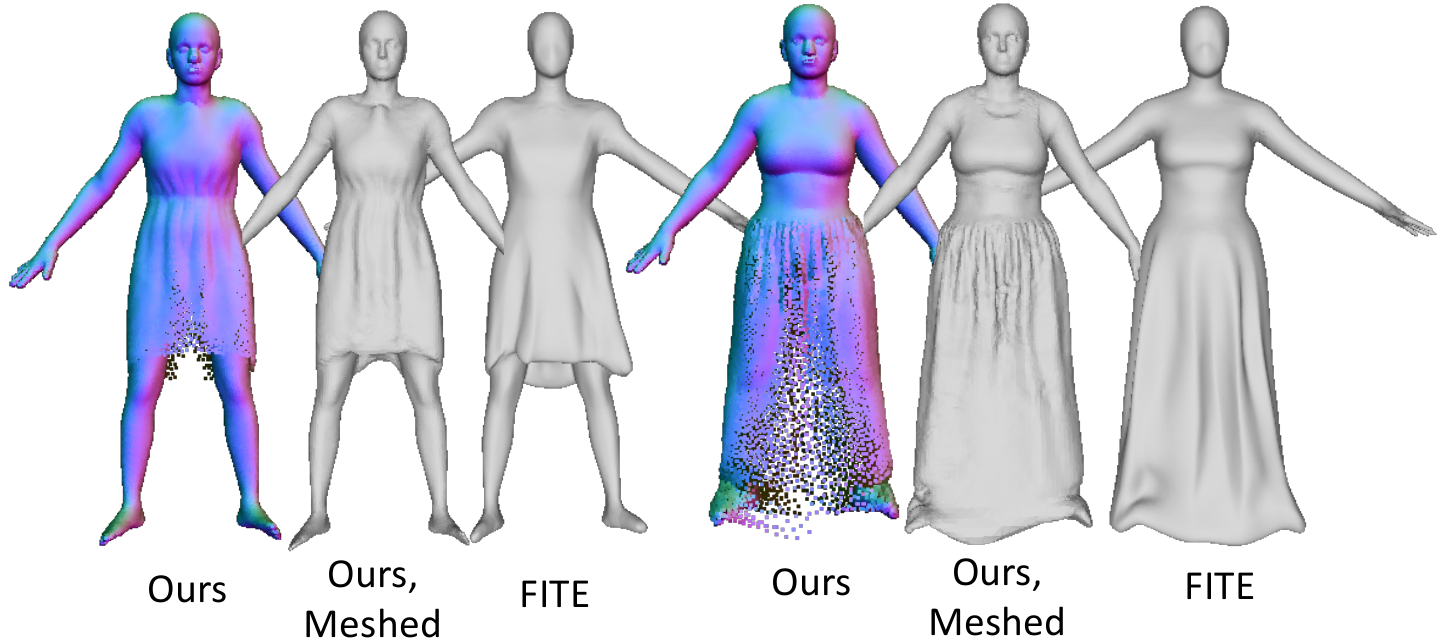}
\vspace{-6mm}
\caption{Comparison of the learned templates with FITE~\cite{lin2022learning}.}
\vspace{-5mm}
\label{fig:tpl_cpr}
\end{figure}

\begin{table}[t]
  \centering
  \caption{Ablation study on the effectiveness of continuous surface features (CSF) and Explicit Template Decomposition (ETD) on a dress outfit (felice-004 from ReSynth~\cite{ma2021power}).}
  \vspace{-2mm}
    \begin{tabular}{lcccc}
    \toprule
    Method & POP  & POP + ETD & CSF  & CSF + ETD \\
    \midrule
    CD & 7.34 & 7.05 & 6.53 & \textbf{6.01} \\
    NML & 1.24 & 1.17 &	1.16 & \textbf{1.16} \\
    \bottomrule
    \end{tabular}%
    \vspace{-5mm}
  \label{tab:ablation_resyn}%
\end{table}%

\paragraph{Evaluation of Explicit Template Decomposition.}
The decomposed templates help to capture more accurate pose-dependent deformations and produce more natural wrinkles in unseen poses, especially for outfits that differ largely from the body template.
To validate the effectiveness of our decomposition strategy, Tab.~\ref{tab:ablation_resyn} reports the ablation experiments on a dress outfit (felice-004) of the ReSynth~\cite{ma2021power} dataset.
We can see that the proposed Explicit Template Decomposition (ETD) brings clear performance gains over baseline methods.
Fig.~\ref{fig:ablation_etd} shows the visual improvements of pose-dependent wrinkles when applying the explicit template decomposition.
Note that the templates are decomposed in an end-to-end manner in our method.
Compared with the implicit template learned in the recent approach FITE~\cite{lin2022learning}, the explicit templates in our method contain more details, as shown in Fig.~\ref{fig:tpl_cpr}.

\section{Conclusions and Future Work}
In this work, we present CloSET, a point-based clothed human modeling method that is built upon a continuous surface and learns to decompose explicit garment templates for better learning of pose-dependent deformations.
By learning features on a continuous surface, our solution gets rid of the seam artifacts in previous state-of-the-art point-based methods~\cite{ma2021scale,ma2021power}.
Moreover, the explicit template decomposition helps to capture more accurate and natural pose-dependent wrinkles.
To facilitate the research in this direction, we also introduce a high-quality real-world scan dataset with diverse outfit styles and accurate body model fitting.

\textbf{Limitations and Future Work.}
Due to the incorrect skinning weight used in our template, the issue of the non-uniform point distribution remains for the skirt and dress outfits.
Combining our method with recent learnable skinning solutions~\cite{saito2021scanimate,ma2022neural} could alleviate this issue and further improve the results.
Currently, our method does not leverage information from adjacent poses. Enforcing temporal consistency and correspondences between adjacent frames would be interesting for future work.
Moreover, incorporating physics-based losses into the learning process like SNUG~\cite{santesteban2022snug} would also be a promising solution to address the artifacts like self-intersections.

\paragraph{\bf Acknowledgements.}
This work was supported by the National Key R\&D Program of China (2022YFF0902200), the National Natural Science Foundation of China (No.62125107 and No.61827805), and the China Postdoctoral Science Foundation (No.2022M721844).

{\small
\bibliographystyle{ieee_fullname}
\bibliography{egbib}
}

\newpage
\appendix

The appendix provides additional details about our approach and more experimental results that were not included in the main manuscript due to limited space.
In Section~\ref{sec:new_dataset}, we present more descriptions of our newly introduced THuman-CloSET dataset.
In Section~\ref{sec:implementation}, we provide more details about the implementation of our approach.
Finally, we report more experimental results in Section~\ref{sec:more_experiments}.
More results are also presented in the Supplementary Video and the project page.


\begin{table}[htbp]
\footnotesize
  \centering
  \caption{Comparison of the scan data used in our experiments.}
    \begin{tabular}{lccc}
    \toprule
    \multirow{2}[1]{*}{Datasets} & \multirow{2}[1]{*}{\# Outfits} & \multirow{2}[1]{*}{Outfit type} & Average \# poses \\
          &       &       & per outfit \\
    \midrule
    CAPE~\cite{ma2020learning}  & 14    & real-world, common & 1806 \\
    ReSynth~\cite{ma2021power} & 12    & synthetic, loose & 984 \\
    THuman-CloSET & 15    & real-world, loose & 140 \\
    \bottomrule
    \end{tabular}%
\vspace{-3mm}
  \label{tab:cape_vs_thuman}%
\end{table}%

\section{THuman-CloSET Dataset}
\label{sec:new_dataset}

\begin{figure}[t]
	\begin{center}
		\includegraphics[width=0.48\textwidth]{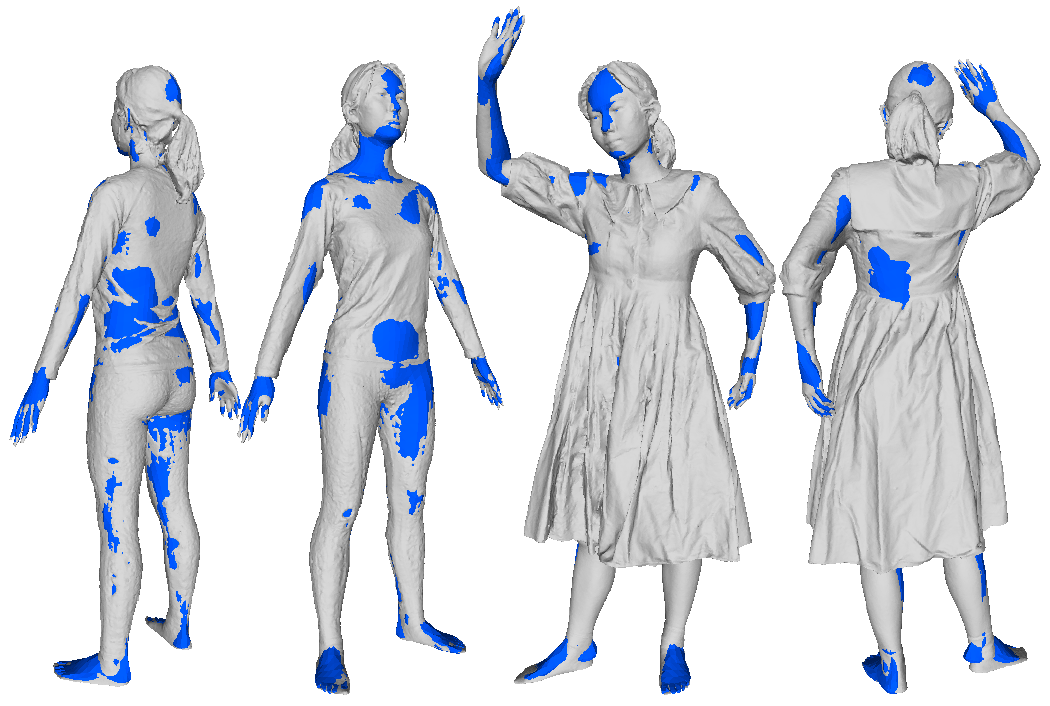}
		\vspace{-5mm}
	\end{center}
    \caption{The fitted SMPL-X models (colored with blue) of the same subject in minimal and loose clothing.}
    \vspace{-5mm}
    \label{fig:smpl_fit}
\end{figure}

\begin{figure*}[t]
    \centering
    \begin{subfigure}[b]{0.53\textwidth}
    \centering
		\includegraphics[width=1.\textwidth]{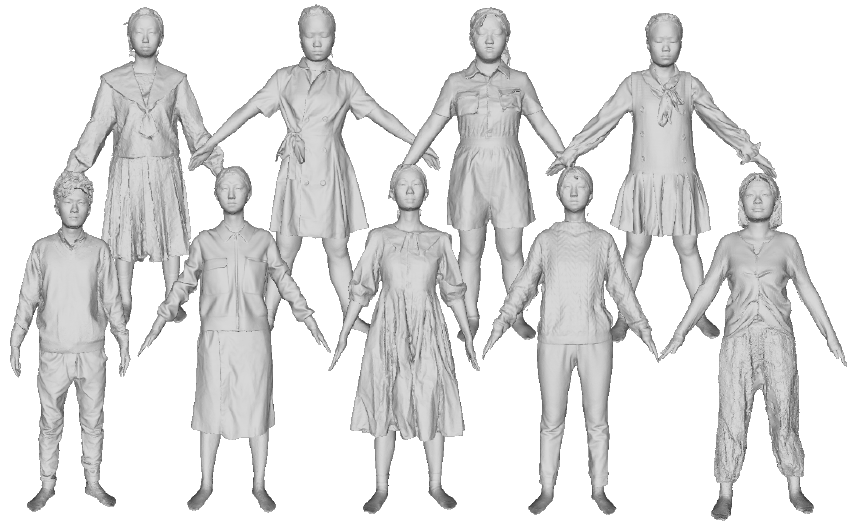}
		\caption{}
		\label{fig:dataset_samples}
    \end{subfigure}
    \begin{subfigure}[b]{0.4\textwidth}
    \centering
		\includegraphics[width=1.\textwidth]{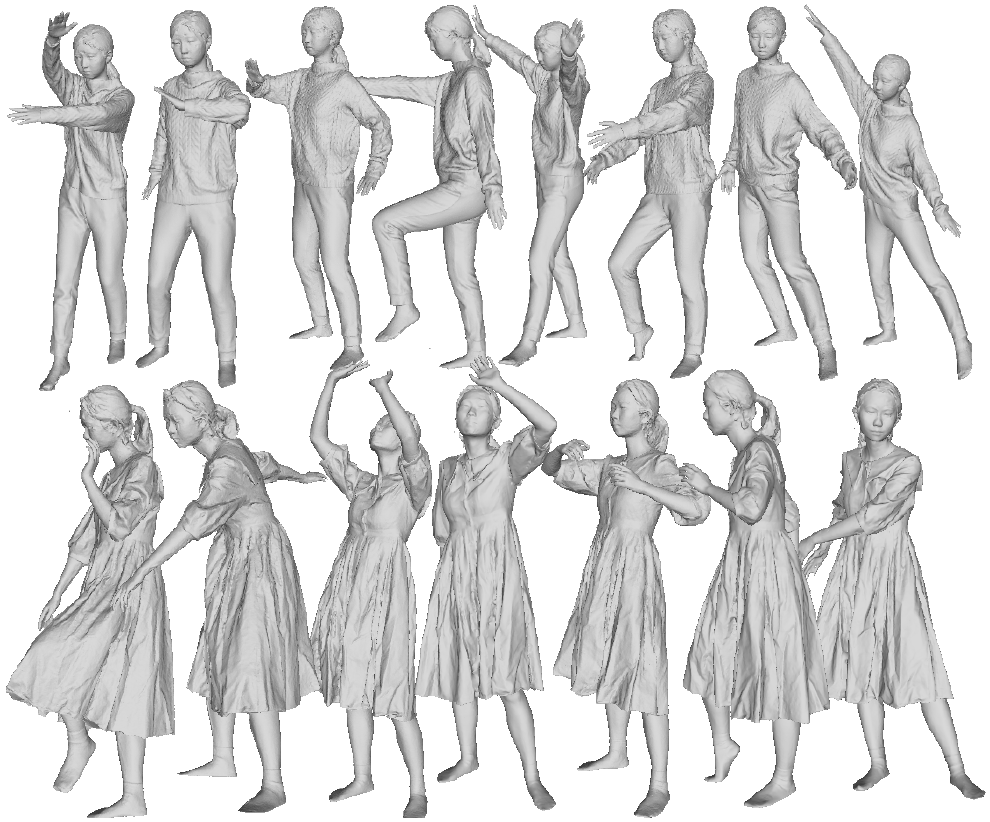}
		\caption{}
		\label{fig:data_pose_sample}
    \end{subfigure}
    \caption{Example scans of our newly introduced THuman-CloSET dataset. (a) Example outfits. (b) Example scans in various poses.}
    \label{fig:data_samples}
\end{figure*}

We introduce THuman-CloSET for the reason that existing pose-dependent clothing datasets~\cite{ma2020learning,ma2021power} are with either relatively tight clothing or synthetic clothing via physics simulation.
THuman-CloSET contains more than 2,000 high-quality human scans captured by a dense camera rig. There are 15 different outfits with a large variation in clothing style, including T-shirts, pants, skirts, dresses, jackets, and coats, to name a few.
All subjects are guided to perform different poses by imitating the poses in CAPE~\cite{ma2020learning}.
For each outfit, there is also a scan of the same subject in minimal clothing so that we can obtain a more accurate body shape.
In our dataset, the body model is firstly estimated from the rendered multiview images of the clothed human and further refined with the ICP optimization between the body model and the scan.
As shown in Fig.~\ref{fig:smpl_fit}, the loose clothing makes the fitting of the underlying body models quite challenging.
For more accurate fitting of the body models, we first fit a SMPL-X~\cite{pavlakos2019expressive} model on the scan of the subject in minimal clothing and then adopt its shape parameters for the fitting of the outfit scans in different poses.
In this way, we ensure that the fitted SMPL-X models of our dataset are overall of good quality.
Fig.~\ref{fig:data_samples} shows several outfit scans and example scans in various poses of THuman-CloSET.
The comparison with CAPE~\cite{ma2020learning}, ReSynth~\cite{ma2021power}, and our THuman-CloSET datasets are summarized in Tab.~\ref{tab:cape_vs_thuman}.
We make THuman-CloSET publicly available for research purposes and hope it can open a promising direction for clothed human modeling and animation from real-world scans.

\section{More Implementation Details}
\label{sec:implementation}
\paragraph{Training.}
Following POP~\cite{ma2021power}, we train our network for 400 epochs on ReSynth~\cite{ma2021power} and CAPE~\cite{ma2020learning} datasets, using the Adam~\cite{kingma2014adam} optimizer with a batch size of 4 and a learning rate of $3.0\times 10^{-4}$.
The loss weights are set to $\lambda_{p}=2\times 10^{4}$, $\lambda_{n}=0.1$, $\lambda_{rgl}=2\times 10^{3}$, $\lambda_{pd}=1.0$, and $\lambda_{gc}=5\times 10^{-4}$ to balance different loss terms.
Note that the normal loss is turned on from the 250th epoch for more stable training, as suggested in~\cite{ma2021power}.

\paragraph{Architecture.}
In the implementation of our network, the PointNet++~\cite{qi2017pointnet++} abstracts the point features for $L=6$ levels, and the numbers of the abstracted points are $2048, 1024, 512, 256, 128, \text{and~} 64$, respectively at each level.
The pose-dependent and garment-related features have the same length of 64, \ie, $C_p=C_g=64$. The decoders $\mathcal{D}_g$ and $\mathcal{D}_p$ adopt the same architecture as POP~\cite{ma2021power}. 
Tab.~\ref{tab:runtime} reports the network parameters and runtime of POP~\cite{ma2021power} and our method.
Note that the pose and garment encoders in our method can also be replaced with recent state-of-the-art point-based encoders such as PointMLP~\cite{ma2022rethinking} and PointNeXt~\cite{qian2022pointnext}.

\begin{table}[htbp]
\footnotesize
  \centering
 \caption{Comparison of the network parameters and runtime.}
    \begin{tabular}{lccc}
    \toprule
    Method & Encoder & \# Params & Runtime \\
    \midrule
    POP~\cite{ma2021power} & UNet~\cite{ronneberger2015u}  & 11.33 M & 44 ms \\
    Ours & PointNet++~\cite{qi2017pointnet++} & 11.76 M & 47 ms \\
    \bottomrule
    \end{tabular}%
\vspace{-2mm}
  \label{tab:runtime}%
\end{table}%

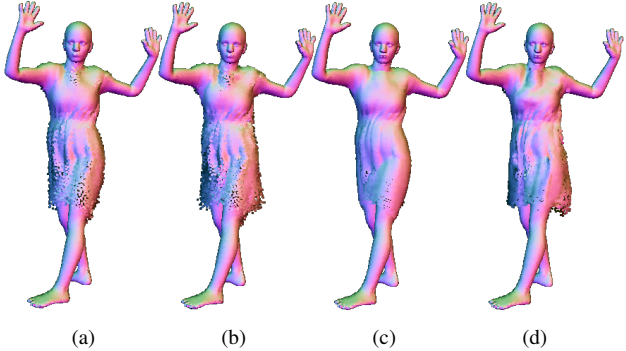
\begin{figure}[t]
	\centering
	\includegraphics[width=0.48\textwidth]{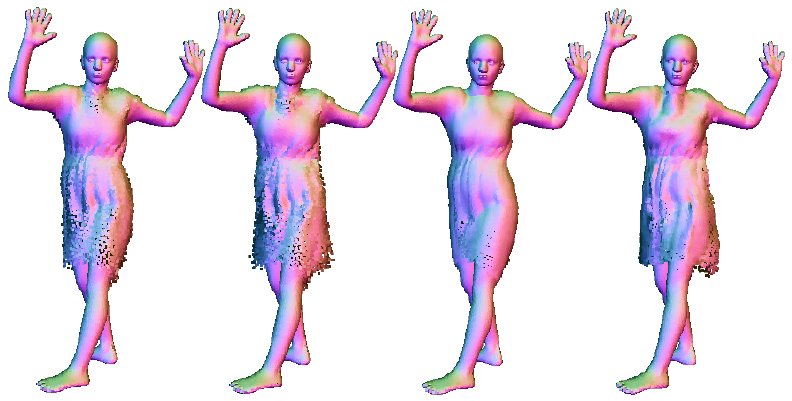}
    \begin{tikzpicture}[remember picture,overlay]
	\node[font=\fontsize{8pt}{8pt}\selectfont] at (-3,0.1) {(a)};
	\node[font=\fontsize{8pt}{8pt}\selectfont] at (-1,0.1) {(b)};
 	\node[font=\fontsize{8pt}{8pt}\selectfont] at (1,0.1) {(c)};
	\node[font=\fontsize{8pt}{8pt}\selectfont] at (3.,0.1) {(d)};
\end{tikzpicture}
\vspace{-2mm}
    \caption{Ablation results on the usage of garment features in the pose decoder. (a)(b) The temple and clothing deformation results without using garment features. (c)(d) The temple and clothing deformation results with the usage of garment features.}
    \label{fig:abla_garment_code}
\end{figure}

\paragraph{Garment Code.}
Following POP~\cite{ma2021power}, for a specific outfit (\eg, an individual garment), the garment code is randomly initialized with the shape of $N\times 64$ ($N$ is the vertex number of SMPL(-X)) and shared across all poses.
During training, the code is optimized with the back-propagated gradients.
When trained with multiple outfits, the pose-dependent deformation should be aware of the outfit type. Hence, the pose decoder takes as input both the garment features $\bm{\phi}_g(\bm{p}^t_i)$ and the pose features $\bm{\phi}_p(\bm{p}^t_i)$.
As shown in Fig.~\ref{fig:abla_garment_code}, the qualitative results become worse when the garment features are not fed into the pose decoder under the multi-outfit setting.

\begin{figure}[t]
    \centering
    \begin{subfigure}[b]{0.2\textwidth}
    \centering
		\includegraphics[height=50mm]{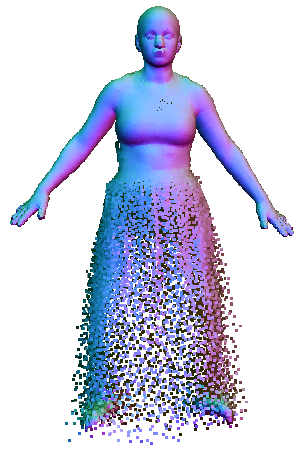}
		\caption{using data term}
		\label{fig:tpl_data}
    \end{subfigure}
    \begin{subfigure}[b]{0.2\textwidth}
    \centering
		\includegraphics[height=50mm]{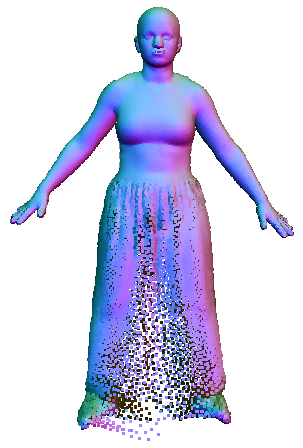}
		\caption{using regularization term}
		\label{fig:tpl_rgl}
    \end{subfigure}
    \caption{The templates learned with (a) the data term and (b) the regularization term.}
    \label{fig:cpr_tpl_ablation}
\end{figure}

\begin{figure*}[t]
	\begin{center}
		\includegraphics[width=0.9\textwidth]{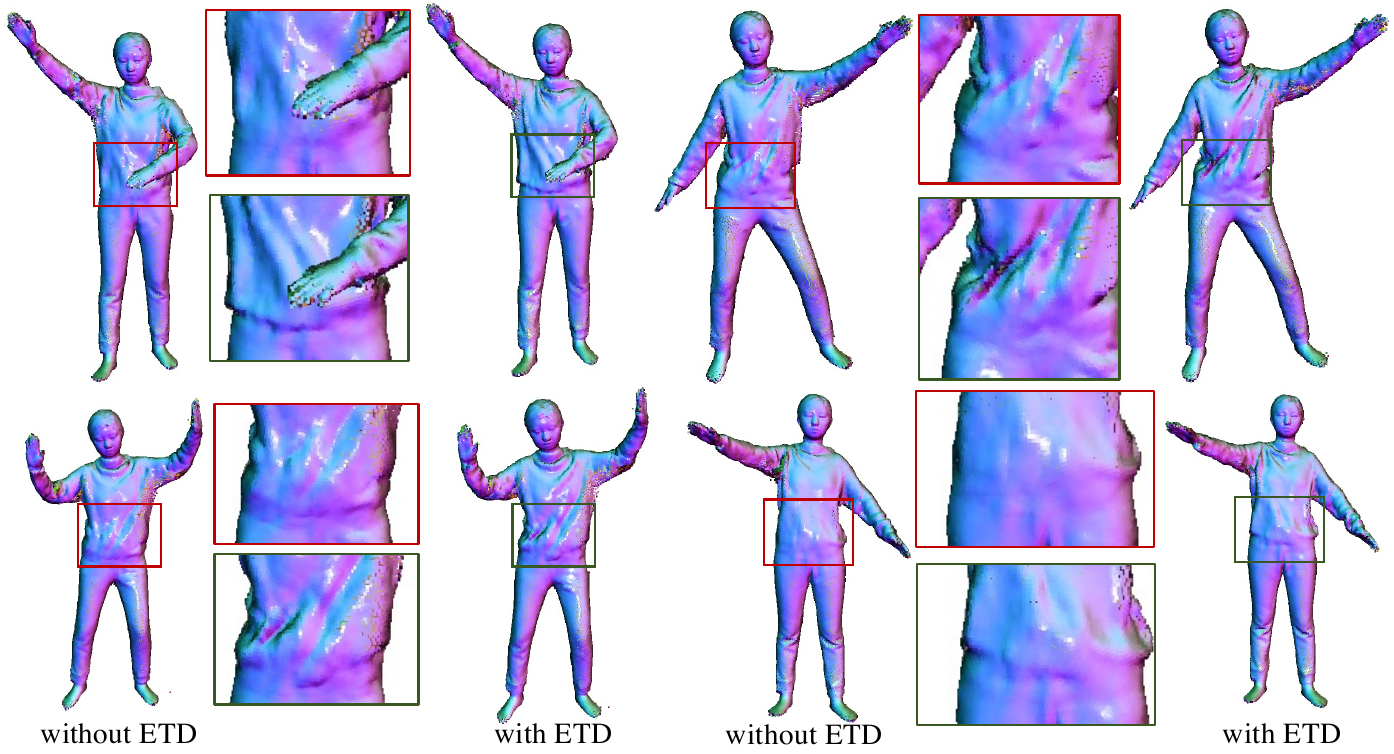}
		\vspace{-5mm}
	\end{center}
    \caption{Comparison of the pose-dependent deformations learned with and without Explicit Template Decomposition (ETD).}
    \label{fig:more_etd_ablation}
\end{figure*}

\section{More Experimental Results.}
\label{sec:more_experiments}

\paragraph{Template learning.}
As described in Section 3 in the main paper, the explicit templates are learned under the regularization term.
An alternative strategy for template learning is applying the data term directly to the generated point clouds of templates, as done in previous work~\cite{ma2022neural}.
However, we found such a strategy leads to worse template learning.
As visualized in Fig.~\ref{fig:cpr_tpl_ablation}, the template directly learned with the data term is much nosier than the one learned with the regularization term.

\begin{table}[t]
  \centering
  \caption{Ablation study of the efficacy of the explicit template decomposition on different backbones. $\dagger$ denotes the default PointNet~\cite{qi2017pointnet} and PointNet++~\cite{qi2017pointnet++}.}
  \label{tab:supp_etd}%
    \begin{tabular}{rlccc}
    \hline
          & Backbone & Size(M) & w/o ETD & w. ETD \\
    \hline
    \begin{sideways}\footnotesize{UV}\end{sideways} & Unet  & 11.33 & 7.34  & 7.05 \\
    \hline
    \multirow{3}[2]{*}{\begin{sideways}Surface\end{sideways}} & PointNet $\dagger$ & 7.68  & 7.14  & 6.94 \\
          & PointNet++ $\dagger$ & 4.35  & 7.08  & 6.71 \\
          & PointNet++  & 11.76 & 6.53  & 6.01 \\
    \hline
    \end{tabular}%
\end{table}%

\paragraph{Effect of Explicit Template Decomposition.}
In Table~\ref{tab:supp_etd}, we further augment the ablation experiments with the backbones of the default PointNet~\cite{qi2017pointnet} and PointNet++~\cite{qi2017pointnet++}.
We can see that i) learning continuous surface features consistently brings improvements over the UV features, though the default PointNet and PointNet++ have smaller model sizes;
ii) PointNet++ is more suitable for surface feature learning than PointNet;
iii) ETD consistently improves the results for all backbones.
In Fig.~\ref{fig:more_etd_ablation}, we include more rendered results of the clothing deformations learned with and without Explicit Template Decomposition (ETD).
In general cases, ETD helps to capture more natural pose-dependent wrinkles.
For more qualitative comparisons of SCANimate~\cite{saito2021scanimate}, SNARF~\cite{chen2021snarf}, POP~\cite{ma2021power}, and our approach, please refer to the supplementary video.

\end{document}